\documentclass[twocol]{ametsocV6.1}
\bibpunct{(}{)}{;}{a}{}{,}

\usepackage{color}

\newcommand{\maybe}{\text{\Large$\color{black}\thicksim$}}
\newcommand{\greencheck}{\text{\Large$\color{black}\checkmark$}}

\title{Comparing Explanation Methods for Traditional Machine Learning Models Part 1: An Overview of Current Methods and Quantifying Their Disagreement}

\authors{Montgomery L. Flora\aff{a,b,e}\correspondingauthor{Montgomery Flora, monte.flora@noaa.gov}, 
Corey K. Potvin,\aff{b,c,e}, 
Amy McGovern\aff{c,d,e}
Shawn Handler\aff{a,b,*}\thanks{*Current Affiliation: Verisk Analytics Incorporated, Jersey City, New Jersey},
}

\affiliation{\aff{a}{Cooperative Institute for Severe and High-Impact Weather Research and Operations, University of Oklahoma, Norman, Oklahoma}\\
\aff{b}{NOAA/OAR/National Severe Storms Laboratory, Norman, Oklahoma}\\
\aff{c}{School of Meteorology, University of Oklahoma, Norman, Oklahoma}\\
\aff{d}{School of Computer Science, University of Oklahoma, Norman, Oklahoma}\\
\aff{e}{NSF AI Institute for Research on Trustworthy AI in Weather, Climate, and Coastal Oceanography}\\
}

\abstract{With increasing interest in explaining machine learning (ML) models, the first part of this two-part study synthesizes recent research on methods for explaining global and local aspects of ML models. This study distinguishes explainability from interpretability, local from global explainability, and feature importance versus feature relevance. We demonstrate and visualize different explanation methods, how to interpret them, and provide a complete Python package (scikit-explain) to allow future researchers to explore these products. We also highlight the frequent disagreement between explanation methods for feature rankings and feature effects and provide practical advice for dealing with these disagreements. We used ML models developed for severe weather prediction and sub-freezing road surface temperature prediction to generalize the behavior of the different explanation methods. For feature rankings, there is substantially more agreement on the set of top features (e.g., on average, two methods agree on 6 of the top 10 features) than on specific rankings (on average, two methods only agree on the ranks of 2-3 features in the set of top 10 features). On the other hand, two feature effect curves from different methods are in high agreement as long as the phase space is well sampled. Finally, a lesser-known method, tree interpreter, was found comparable to SHAP for feature effects, and with the widespread use of random forests in geosciences and computational ease of tree interpreter, we recommend it be explored in future research. 
}

\begin{document}
\maketitle

\statement{The primary goal of this paper is to make the atmospheric science community aware of recently developed explainability methods for traditional ML models and demonstrate their use in a software package developed by the authors (scikit-explain; \citealt{Flora+Handler}). Another important goal is to highlight that these methods tend to disagree with each other, which can affect the trust in these products. Like forecast verification metrics that describe different aspects of forecast quality, some disagreement is to be expected as explainability methods are designed for different tasks. Finally, we offer some practical advice on explaining the behavior of ML models when different explanation methods disagree with each other. We hope that this will motivate future work to explore the quality of explainability methods. For example, in Part II of this study, we find that knowing the relative faithfulness of feature ranking methods can objectively improve explainability.}

\section{Introduction}\label{intro}
Machine learning algorithms (ML) are increasingly common in severe weather applications (e.g., Gagne et al. \citeyear{Gagne+etal2017}; \citealt{Lagerquist+etal2017, Cintineo+etal2020, Lagerquist+etal2020, Flora+etal2021}), ensemble post-processing (e.g., \citealt{Rasp+Lerch2018}), sub-freezing road temperature prediction \citep{Handler+etal2020}, model parameterization (e.g., \citealt{Brenowitz+etal2020}), tropical cyclone prediction (e.g., \citealt{Kumler-Bonfanti+etal2020}), and climate modeling (e.g., \citealt{Hernadez+etal2020}). A key advantage of ML models is their ability to leverage multiple input features and learn useful multivariate relationships for prediction, calibration, and post-processing.  However, many ML models are considered "black boxes" in that the end-user cannot understand the internal workings of the model \citep{McGovern+etal2019_blackbox}. ML systems in low-risk situations may not need to be understood, but in high-risk situations\textemdash like severe weather forecasting\textemdash decision makers want to know why a model came to its prediction. As part of building trust with human forecasters in ML predictions, it is important to explain the "why" of an ML model's prediction in understandable terms and to create real-time visualizations of these methods (\citealt{Hoffman+etal2017, Karstens+etal2018, Jacovi+etal2021}).

\subsection{Interpretability vs. Explainability}

\begin{table*}[h]
\caption{ A non-exhaustive list of definitions of interpretability and explainability provided in the literature. Many studies not included here do not define the terms and use them interchangeably. These are partial quotes from each source, but for readability, quotation marks are omitted.} \label{table:definitions}
\begin{center}
\begin{tabular}{|p{2.75cm}||p{6cm}||p{6cm}|}

\hline \hline
Source & \textbf{Interpretability} & \textbf{Explainability} \\ 
\hline

\citet{Kim+etal2016} & [\ldots] a method is interpretable if a user can correctly and efficiently predict the method's results. & N/A (no distinction made) \\ 

\citet{Doshi-Velez+Kim2017} & [\ldots] the ability to explain or to present [the model] in understandable terms to a human. &  explaining the model after it is trained with post-hoc methods \\

\citet{Adadi+Berrada2018} & [\ldots] interpretable systems are explainable if their operations can be understood by human[s]. & N/A (no explicit definition provided, but the terms are not treated interchangeably)  \\

\citet{Rudin+etal2018} & An interpretable machine learning model is constrained in model form so that it is either useful to someone, or obeys structural knowledge of the domain such as [\ldots] the physical constraints that come from domain knowledge. &  [\ldots] where a second (post-hoc) model is created to explain the black box model \\ 

\citet{Gilpin+etal2018} & [\ldots] describe the internals of a system in a way which is understandable to humans & [\ldots] models that are able summarize the reasons for [black box] behavior [...] or produce insights about the causes of their decisions \\

\citet{Murdoch+etal2019} & [\ldots] the use of machine-learning models for the extraction of relevant knowledge about domain relationships contained in data & N/A (no distinction made) \\ 

\citet{Miller+etal2019} & the degree to which a human can understand the cause of a decision. & N/A (no distinction is made) \\

\citet{Lindardatos+etal2020} & [ability] to identify cause-and-effect relationships within the system’s inputs and outputs. & Explainability, [...] is associated with the internal logic and mechanics that are inside a machine learning system. \\

\citet{InterpretMLTextbook} & Adopts the definitions from \citet{Miller+etal2019} and \citet{Kim+etal2016} & N/A (no distinction is made; instead distinguishes interpretability/explainability from \textit{explanation} where explanation refers to explaining individual predictions). \\

\citet{Rudin+etal2021} & An interpretable ML model obeys a domain-specific set of constraints to allow it to be more easily understood by humans. These constraints can differ dramatically depending on the domain. & Explaining a black box model with a simpler model \\

Wikipedia & describes the possibility to comprehend the ML model and to present the underlying basis for decision-making in a way that is understandable to humans. & the collection of features of the interpretable domain, that have contributed for a given example to produce a decision  \\ 

\hline
\end{tabular}
\end{center}
\end{table*}

Many methods have been developed to better understand black box models, and in response, substantial research has emerged on topics such as \textit{interpretable ML} and \textit{explainable artificial intelligence} (XAI) (e.g., \citealt{vanLent+etal2004, Kim+etal2016, Adadi+Berrada2018, Rudin+etal2018, Gilpin+etal2018, Miller+etal2019,  Lindardatos+etal2020,  Molnar+etal2020_imlpaper, Rudin+etal2021}). Given the nascency of these topics, the definitions of \textit{explainability} and \textit{interpretability} are inconsistent throughout the literature, and many articles treat them interchangeably (Table~\ref{table:definitions}). In this paper, we define these terms as follows:
\begin{itemize}
    \item \textit{Interpretability} is the degree to which an entire model and its components are capable of being understood without additional methods; and
    \item \textit{Explainability} is the degree to which any partially interpretable or uninterpretable model (i.e., black boxes) can be approximately understood through post hoc methods (e.g., verification, visualizations of important features or learned relationships). 
\end{itemize}
This distinction between interpretability and explainability is needed, since some in the ML and statistics community favor producing interpretable models (i.e., restricting model complexity beforehand to impose interpretability; \citealt{Rudin+etal2018, Rudin+etal2021}), while the general trend in the ML community is to continue developing partially interpretable and black box models and implementing post hoc methods to explain them. \citet{Lipton2016} defines a fully interpretable model as one that has \textit{simulatability} (the entire model can be considered at once), \textit{decomposability} (each component of the model is human-understable) and \textit{algorithmic transparency} (one can understand how the model was trained). A partially interpretable model may only meet one of these criteria. Explainability can be further subdivided into \textit{model-specific explainability} (where the components of the model can be used for the explanation) and \textit{}model-agnostic explainability (where no components of the model itself are used and no assumption is made about the model structure).

\begin{figure*}[t]
  \noindent\includegraphics[width=38pc,angle=0]{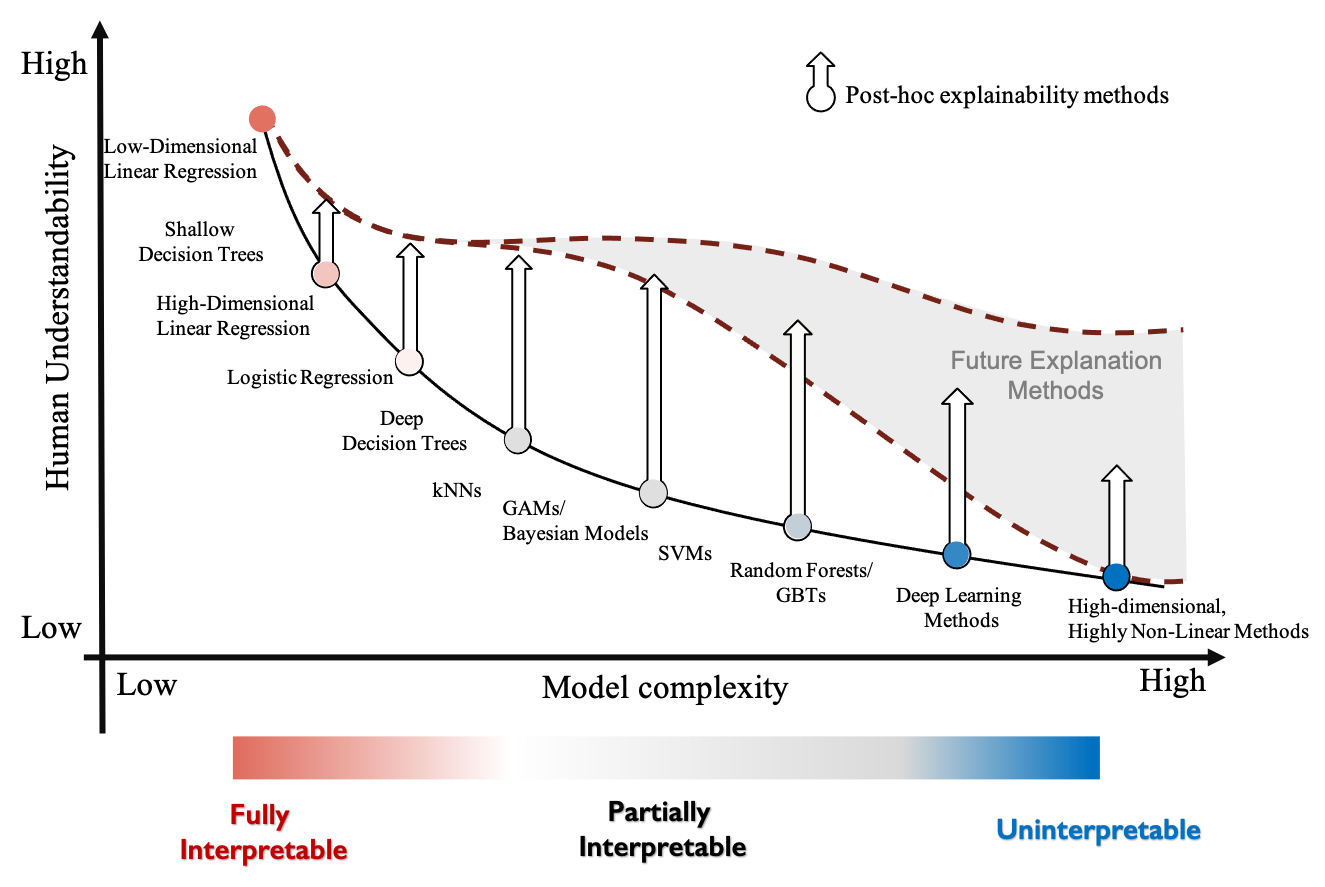}\\
  \caption{Illustration of the relationship between understandability and model complexity. Fully interpretable models have high intrinsic understandability while partially interpretable or simpler black box models have the most to gain from explainability methods. With increased dimensionality and non-linearity, explainability methods can improve understanding, but there is considerable uncertainty about the ability of future explanation methods to improve understandability of high-dimensional, highly non-linear methods. }\label{fig:int_vs._exp}
\end{figure*}

An illustration of interpretability and explainability is provided in Figure ~\ref{fig:int_vs._exp}. Fully interpretable models do not require post hoc explainability methods to improve understanding, while uninterpretable models have the most to gain from additional explanation methods. For example, low-dimensional linear regression is fully interpretable and a shallow decision tree is partially interpretable. In contrast, a deep neural network (DNN) or a dense random forest is uninterpretable but, through external explanation methods, can be approximately understood. Explanation methods must be approximate, as they would otherwise be as incomprehensible as the black-box model itself. We do not view this as a limitation of explanation methods, as suggested by other studies (e.g. \citealt{Rudin+etal2018, Rudin+etal2021}), since abstracting a complex model is required for human understanding. For example, it is common in the weather community to replace the full Navier-Stokes equation with conceptual models that are more understandable. However, the degree of explanability is controlled by the complexity of the model \citep{Molnar+etal2019}. As the number of features increases or their interactions become more complex, the explanations for the behavior of the ML model become less compact and possibly less accurate. At the moment, it is unclear how much improvement in our understanding of high-dimensional, highly nonlinear models current and future explanation methods will offer (Fig.~\ref{fig:int_vs._exp}). 

\subsection{Explainability Methods and their Disagreements}
The atmospheric science community is beginning to adopt explainability methods (e.g., \citealt{Lakshmanan+etal2015, Minokhin+etal2017, Herman+Schumacher2018a, Rasp+Lerch2018, McGovern+etal2019_blackbox, Jergensen+etal2020, Lagerquist+etal2020,  Gagne+etal2019, Handler+etal2020, Hamidi+etal2020, Mecikalski+etal2021, Loken+etal2022, Shield+Houston2022}). Given the increasing interest in model explainability, in this first part of a two-part study, we synthesize recent research on multiple explainability methods using ML models developed for severe weather prediction \citep{Flora+etal2021} and sub-freezing road surface temperature prediction \citep{Handler+etal2020}. By highlighting multiple explainability methods, our goal is to also demonstrate the degree to which explanations of model behavior can disagree with each other. Only recently have studies begun to document the sometimes substantial disagreement between explainability methods (e.g., \citealt{McGovern+etal2019_blackbox, Molnar+etal2020_imlpaper, Satyapriya+etal2022_disagree, Covert+etal2020}).  \citet{McGovern+etal2019_blackbox} is a seminal text for introducing multiple traditional ML and deep learning explainability methods to the atmospheric science community, but the analysis of disagreement among methods was largely subjective. \citet{Satyapriya+etal2022_disagree} is one of the first studies to quantify the disagreement of different explainability outputs, but their study was limited to local attribution methods and it is difficult to determine how their results translate to highly complex datasets commonly found in atmospheric sciences.  

Although previous studies have highlighted explanation disagreement (e.g., \citealt{McGovern+etal2019_blackbox, Satyapriya+etal2022_disagree}), a thorough and objective exploration of explanation disagreement for a variety of explainability methods and in the context of meteorological data is warranted. We adopt methods similar to those of \citet{Satyapriya+etal2022_disagree} to measure the global feature ranking agreement and introduce similar metrics to measure the agreement of estimated feature effects. In \citet{PartII} (hereafter Part II), we perform experiments similar to those in \citet{Covert+etal2020} to measure the validity of feature ranking methods. By interrogating and measuring the disagreement between explainability methods, we hope to motivate further research on explanation methods that evaluate properties such as those introduced in \citet{OpenXAI} (e.g., faithfulness, stability, and fairness).

Our contributions include highlighting the distinction between \textit{intepretability} and \textit{explainability} (discussed above), summarizing the various explainability methods and their key ideas (Section \ref{sec:explain_methods}), suggestions for interpreting the different explainability output, and providing practical advice for explaining ML models when explanations seemingly disagree with each other. Lastly, we introduce a complete Python package (scikit-explain; \citealt{Flora+Handler}) that computes and visualizes all the explainability methods demonstrated here (and more not shown).

The structure of this paper is as follows. Sections \ref{data} and \ref{method} describe the severe weather and road surface datasets and the ML models used, respectively. Section \ref{sec:explain_methods} covers the global and local explanation methods explored in this study. We present the results in Section ~\ref{sec:results} and in Section ~\ref{sec:advice} we provide some practical advice to explain ML models, particularly when the explanations from different methods disagree. Finally, the conclusions and limitations of the study are discussed in Section ~\ref{sec:conclusions}. 

\section{ Data } \label{data} 
\subsection{Description of the Severe Weather Dataset}\label{sec:wofs_data}
The severe weather data set is derived from the output of the 2017-2019 Warn-on-Forecast System (WoFS), which is an experimental 3-km ensemble that produces rapidly updating forecast guidance at 0-6-h lead times.  Additional details of the WoFS are found in \citet{Wheatley+etal2015}, \citet{Jones+etal2016}, and \citet{Jones+etal2020}. The ML dataset contains features derived from intrastorm and environmental variables extracted from within 30-min ensemble storm tracks \citep{Flora+etal2019,Flora+etal2021} with additional features that describe the morphological attributes of the tracks themselves (e.g., minor and major axis length; see Table 1 from \citealt{Flora+etal2021}). Environmental features are spatial averages (within a track) of the ensemble mean and standard deviation fields valid at the beginning of the 30-min forecast period. Intrastorm features include both the spatial average values of ensemble mean and standard deviation fields (similar to environmental features) from time-composited fields and the ensemble mean and standard deviation of spatial 90th percentile values of each ensemble member within a storm track (meant to capture storm intensity). The same ML features are used for each severe weather hazard prediction model, but the target variable is whether the ensemble storm track is matched to a tornado, severe hail, or severe wind report, respectively.  Although the same features are used for all hazards, the dates from which the features were computed varied to help ensure similarity between the training and testing datasets \citep{Flora+etal2021}.

\subsection{Description of the Road Surface Dataset}
The road surface dataset from \citet{Handler+etal2020} spans two cool seasons: 1 October 2016 – 31 March 2017 and 1 October 2017 – 31 March 2018. Thirty features were used for training, including near-surface variables from the High Resolution Rapid Refresh (HRRR) model as well as derived features (\citealt{Handler+etal2020}, Table 1). The variables were informed by previous research that identified variables important for modulating surface road temperatures (e.g., \citealt{Crevier_2001}). Hourly road surface temperature observations from the Road Weather Information System (RWIS) sites were used as the target variable for training. Each example was labeled below or above freezing based on the temperature reported by the RWIS site. The RWIS sites used are shown in Fig. 1a of \citet{Handler+etal2020}. 

\section{Machine Learning Algorithms} \label{method} 
For this study, we used classification random forests and logistic regression models available in the Python sci-kit learn package \citep{Pedregosa+etal2011}. Consistent with \citet{Handler+etal2020} and \citet{Flora+etal2021}, the random forest is applied to the road surface dataset to predict whether a road will freeze while logistic regression is applied to the WoFS dataset to predict whether a storm track will be associated with a severe hail, severe wind, or tornado report, respectively. 

\subsection{Logistic Regression with Elastic Nets} \label{sec:log_reg}
A logistic regression model is a linear regression model designed for classification tasks. Given a binary outcome variable $y$ (1 or 0), we can estimate the probability that $y$ belongs to a particular class (e.g, $P(y=1 | X)$) as:
\begin{equation}\label{eqn:log_reg}
    P(y=1 | X) = \frac{1}{1 + \exp(-\beta_0 - \sum_{i=1}^N \beta_ix_i)}
\end{equation}
where $\beta_i$ are the learned weights, $x_i$ are the features and $\beta_0$ is the bias term. Although logistic regression is based on a linear model, the predicted probability is not linearly related to the input features. In fact, unless the features are binary, binary classification is inherently non-linear due to the non-linear transformation of continuous features into binary target variables ($\mathbb{R}^{m\times n} \rightarrow \{0,1\}^m$ where $m$ and $n$ are the number of examples and features, respectively). Given that the summation in eqn. \ref{eqn:log_reg} is an exponent, there is a multiplicative interaction between all $N$ features. Therefore, the logistic regression model has questionable interpretability in probability space, especially as the number of features increases. In this study, we treat logistic regression as partially interpretable and explain it using both its coefficients and external explanation methods. Regularizations, both L1 and L2, are used for training. L1 regularization acts as a feature selection method by zeroing coefficients for bad features while L2 regularization encourages smaller weights and thereby discourages the model for heavily favoring features. 

\subsection{Random Forest} \label{sec:rf}
An increasingly popular ML algorithm is the random forest \citep{Breiman2001_RF}. A classification random forest is made up of multiple decision trees, each partitioning the feature space into subregions of increasing "purity" (homogeneity of the target variable). To improve the predictive accuracy of the random forest, each tree is trained on a bootstrapped resampled version of the data, and for each split only a small random subset of features is considered. For each tree, the prediction is the proportion of positive class examples (number of positive class examples divided by the total number of examples in the leaf node). The final prediction of the forest is the ensemble average of the separate tree predictions. The random forest in this study is treated as uninterpretable and we explain it using model-specific and model-agnostic explainability methods. 


\begin{figure*}[ht]
  \noindent\includegraphics[height=46pc, width=36pc,angle=0]{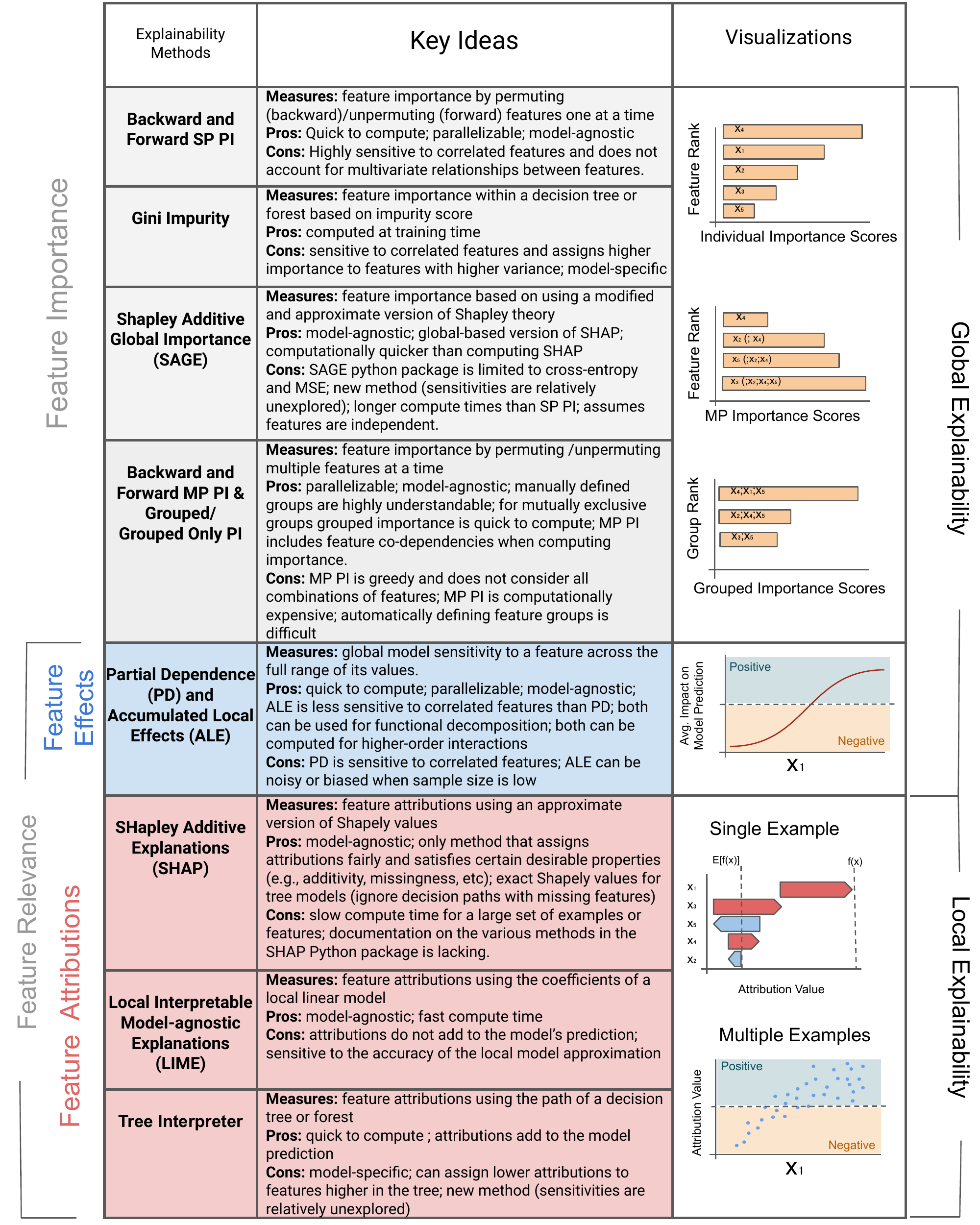}\\
  \caption{Explainability methods covered in this study, their key ideas and typical visualizations. SP and MP are short for single-pass and multi-pass, respectively. PI is short for permutation importance.
  }\label{fig:summary}
\end{figure*}

\section{Explainability Methods}\label{sec:explain_methods}
    In line with \citet{Lipton2016}, \citet{InterpretMLTextbook} identifies five scopes of ML explainability, which can be summarized into three main categories:
    \begin{itemize}
        \item \textit{Algorithmic Transparency}: How does the algorithm create the model?
        \item \textit{Global Explainability}: How does the trained model as a whole make predictions? How do parts of the model affect the predictions?
        \item \textit{Local Explainability}: Why did the model make a certain prediction for a single example? Why did the model make specific predictions for a group of examples?
    \end{itemize}
    Model explanability typically refers to global or local explainability, whereas algorithmic transparency does not refer to a specific model or prediction. Both global and local explainability methods can be summarized as measuring and visualizing: 
    \begin{itemize}
        \item \textit{Feature importance}: the ranking of features or sets of features by how much they contribute to a model's predictions or performance (e.g., \citealt{Breiman2001_RF, Lakshmanan+etal2015, Greenwell+etal2018, Lundberg+Lee2016});
        \item \textit{Feature effects}: the expected functional relationship between a feature (or set of features) and a ML model's prediction (e.g., \citealt{Friedman2001, Apley+etal2016, Greenwell+etal2018, Lundberg+Lee2016}); and
        \item \textit{Feature interactions}: how a given feature's effect is dependent on other features and the strength of that effect (e.g., \citealt{ Friedman+Popescu2008, Greenwell+etal2018, Molnar+etal2019, Oh+etal2019,feature_interact_txtbk}).
    \end{itemize}
    Global approaches attempt to decompose the model into parts that can be understood individually \citep{Murdoch+etal2019,Molnar+etal2020_imlpaper}. Local approaches explain individual predictions. These methods can include, but are not limited to, decomposing a prediction into the contribution of each feature (e.g., \citealt{Ribeiro+etal2016, Lundberg+Lee2016}) or developing counterfactual explanations to form what-if scenarios \citep{Molnar+etal2020_imlpaper, InterpretMLTextbook}. It is important to employ multiple explanation methods, as no one method is "one-size-fits-all" \citep{Molnar+etal2021}.  A brief summary of the key ideas for each of the explainability methods discussed in the following sections can be found in Figure ~\ref{fig:summary}. Furthermore, inspired by \citet{Covert+etal2020}, Table ~\ref{tab:rankings_methods} provides a summary of the properties of the different explainability methods used in this study. 
\begin{table*}[]
    \centering
    \begin{tabular}{lcccccc}
        \hline
        Method                      &  SCALE & AGNOSTIC & PERFORM     & INTERACT       & MISSINGNESS  & TRACTABLE \\
        \hline 
        Linear Model Coeffs.        &  G & $\times$     & $\times$     & $\times$      & $\times$     &  \greencheck \\ 
        Tree Interpreter            &  L & $\times$     & $\times$     & \maybe        &  \greencheck &  \greencheck \\ 
        PD                          &  G & \greencheck  & $\times$     & $\times$      &  $\times$    &  \greencheck   \\ 
        ALE                         &  G & \greencheck  & $\times$     & $\times$      &  \greencheck &  \greencheck   \\ 
        SHAP                        &  L & \greencheck  & $\times$     & \greencheck   &  \maybe      &  $\times$ \\ 
        LIME                        &  L & \greencheck  & $\times$     & $\times$      &  \maybe      &  \greencheck \\ 
        \hline
         Gini Importance             &  G & $\times$     & \greencheck    & $\times$      &  $\times$    &  \greencheck \\ 
        Backward Single-Pass PI    & G & \greencheck  & \greencheck  & $\times$     &  \maybe   &  \greencheck  \\ 
        Forward Single-Pass PI    & G & \greencheck  & \greencheck   & $\times$     &  \greencheck  &  \greencheck  \\ 
        Backward Multipass PI      & G &  \greencheck  & \greencheck  & \maybe        &  \maybe   &  \maybe \\
        Forward MultiPass PI       & G &  \greencheck  & \greencheck   & $\times$      &  \greencheck     &  \maybe \\
        SAGE                       & G & \greencheck  & \greencheck   & \greencheck   &  \maybe   &  \maybe \\
        \hline
    \end{tabular}
    \caption{Comparison of feature importance and relevance explainability methods covered in this paper and their properties. \textit{Scale}: method is global (G) or local (L). \textit{Agnostic}: method works with any model class. \textit{Perform}: scores are related to the performance gains associated with each feature. \textit{Interact}: feature interactions are considered (i.e., individual feature importance/relevance accounts for multivariate relationships). \textit{Missingness}: held out features are accounted for properly (e.g., by training a new model, or marginalizing them out). \textit{Tractable}: method is computationally efficient (e.g., computed at training time, parallelizable, etc.). The symbol \greencheck shows that a property is satisfied, $\times$ that it is not, and \maybe \ that it is partly satisfied. PI is short for permutation importance. }
    \label{tab:rankings_methods}
\end{table*}

\subsection{Global Explainability Methods}\label{sec:global_explain}
    \subsubsection{Feature Importance vs. Relevance}\label{sec:feature_rankings}
        Ranking features within a dataset or based on their contribution to the model is a crucial component of model interpretability and explainability. In the literature, feature ranking methods tend to fall into one of three categories: 
        \begin{enumerate}
            \item Strength of univariate relationship with the target variable
            \item Expected contribution to the magnitude of a model's prediction 
            \item Expected contribution to the model's performance/accuracy
        \end{enumerate}
        The first category does not involve the model itself and reflects characteristics of the data, such as correlations with the target variable or the Kullback-Leibler J measure \citep{Lakshmanan+etal2015}. Regression coefficients, feature attribution methods (e.g., SHAP, tree interpreter, LIME; see Section \ref{sec:explain_methods}\ref{sec:local_explain}), and partial dependence/accumulated local effect variance \citep{Greenwell+etal2018} are examples of the second category, while different variations of permutation importance,  Gini impurity importance \citep{Breiman2001_RF, McGovern+etal2019_blackbox}, Shapley Additive Global importancE (SAGE; \citealt{Covert+etal2020}) or sequential feature selection \citep{McGovern+etal2019_blackbox} are examples of the third category. 
        
        In general, the first two categories can be defined as measures of \textit{feature relevance}, while \textit{feature importance} is formally defined with respect to model performance \citep{vanderLaan+2006, Covert+etal2020}. We can further separate the notion of feature importance into \textit{model-specific feature importance} and \textit{model-agnostic feature importance}. Model-specific importance quantifies how much a set of features contributes to the performance of a given model, while model-agnostic importance quantifies the hypothetical range in which any well-performing model may rely on a set of features for model performance \citep{Fisher+etal2018, Covert+etal2020}\footnote{The idea of model-specific importance and model-agnostic importance is referred to as model-based predictive power and universal predictive power in \citet{Covert+etal2020}. \citet{Fisher+etal2018} loosely refers to the notion of model-agnostic importance with the idea of \textit{model class reliance}: the highest and lowest degree to which any well-performing model with a given class may rely on a predictor for prediction accuracy.}. For example, although sequential feature selection considers model performance, it does not do so with respect to the model trained on the original set of features, and situations can arise where, due to compensatory effects, one seemingly important variable is removed and the model adjusts using the remaining variables \citep{feature_interact_txtbk}. Therefore, sequential feature selection is an approximation of model-agnostic feature importance, while variations on the permutation methods, which do not alter the original model, are forms of model-specific feature importance. 
        
    \subsubsection{Permutation Importance Methods}\label{sec:flavors_of_perm}
        Permutation importance is one of the most popular methods for assessing feature importance. It was first introduced in \citet{Breiman2001_RF}, but was later expanded in \citet{Lakshmanan+etal2015}. Recent studies have referred to the methods in \citet{Breiman2001_RF} and \citet{Lakshmanan+etal2015} as single-pass and multipass permutation importance methods, respectively (e.g., \citealt{McGovern+etal2019_blackbox, Jergensen+etal2020, Handler+etal2020}). The main goal of both methods is to measure the expected model accuracy when the values of a feature are permuted so that the feature becomes uninformative of the target variable, but the marginal distribution of the feature is maintained. If the expected model accuracy is relatively unchanged after the feature values are shuffled, then the feature is considered unimportant. The single-pass method permutes only one feature at a time. For the multipass method, the most important feature from the single-pass is left permuted, and then the remaining features are permuted one at a time to determine the second most important feature. The top 2 features are then left permuted to determine the third most important feature, and so on.  \citet{Lakshmanan+etal2015} developed this technique to emulate sequential backward selection (\citealt{McGovern+etal2019_blackbox}), except that instead of removing a feature and refitting the model, features are left permuted to effectively remove them from the model. Sequential forward selection can also be emulated by jointly permuting all features and then unpermuting them one at a time to determine the most important feature. The most important feature remains unpermuted, and each remaining feature is unpermuted again to determine the second most important feature (and so on). Similarly to the nomenclature for sequential feature selection, we can refer to the latter and former permutation methods as the backward and forward multipass permutation importance\footnote{These definitions of backward and forward are opposite of \citet{Lagerquist+etal2021}, in which their labels do not refer to the type of sequential feature selection, but to the permutation order. For example, with all features permuted, \citet{Lagerquist+etal2021} interprets restoring features (i.e., unpermuting) as moving \textit{backwards} with respect to when the features were permuted.}, respectively. In summary, backward methods measure how removing a feature (or set of features) reduces model performance, while forward methods measure how much the model performance relies solely on the unpermuted features. 

        A lauded advantage of the multipass method over the single-pass method is that by permuting more than one feature, it accounts for feature codependencies \citep{Lakshmanan+etal2015, Gregorutti+etal2015_grouped}. For example, multiple studies have found that when features are correlated, their single-pass permutation importance score can be reduced \citep{Strobl+etal2007, Strobl+etal2008, Gregorutti+etal2015_grouped, Gregorutti+etal2017_corr}. Furthermore, when permuting more than one feature, the total importance is not equal to the sum of their individual importances, as it also depends on the codependencies between the features \citep{Gregorutti+etal2015_grouped}.  Thus, multipass permutation importance is usually interpreted as taking feature codependencies into account. 

\begin{figure}[t]
        \centering
        \noindent\includegraphics[width=3.5in, angle=0]{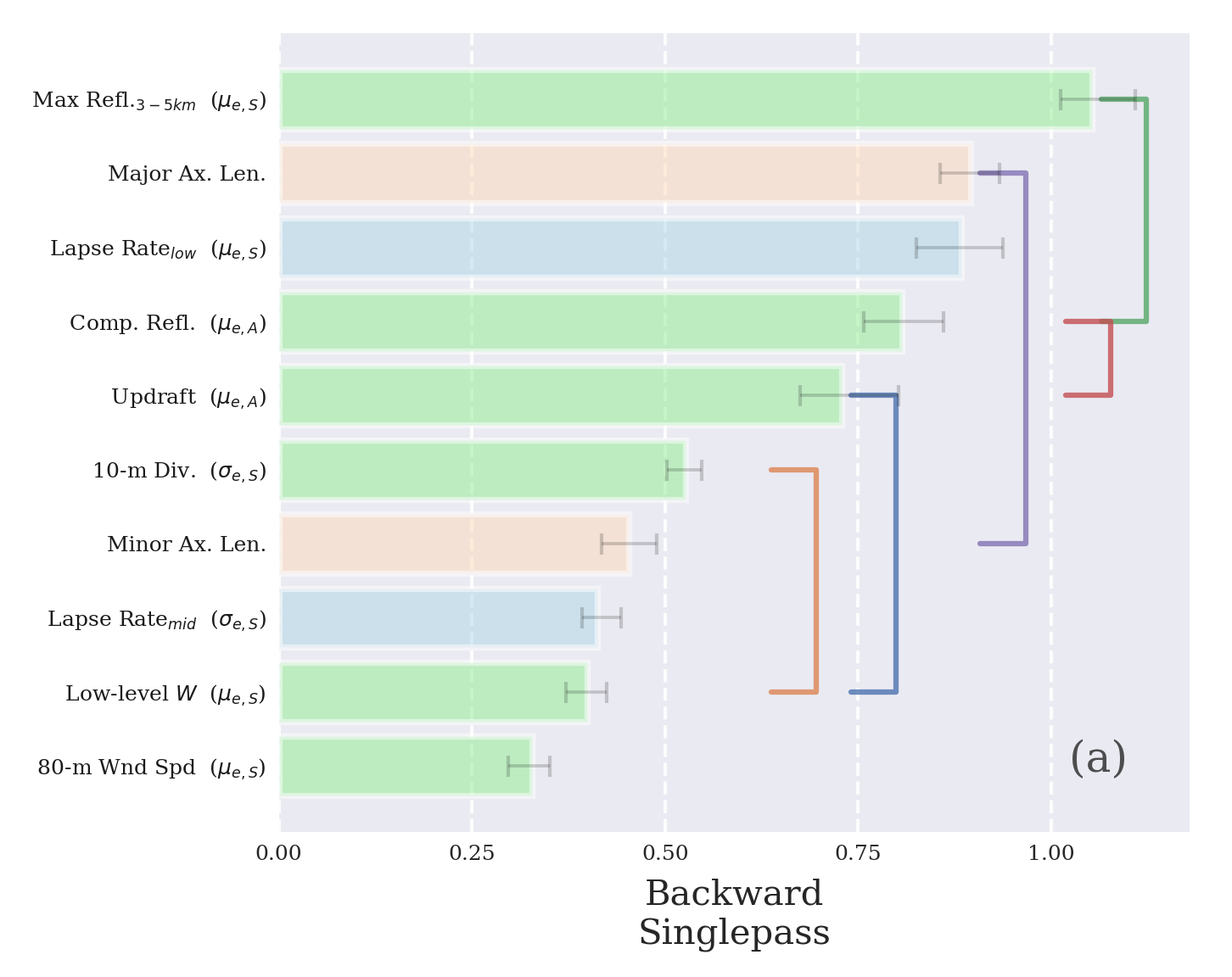}
        \caption{ Backward single-pass permutation importance ranking for the severe hail dataset. Green, blue, and light gold indicate intra-storm, environment, and storm morphological features, respectively. Features with a linear correlation $>$ 0.5 are linked together (each linkage is shown in a different color to help distinguish them from one another). See Table 1 in \citet{Flora+etal2021} for the feature naming convention. }
        \label{fig:hail_rankings}
        \end{figure}

        To summarize, the single-pass permutation importance measures the univariate contribution of features to the model performance, while the multipass permutation importance attempts to measure the multivariate contribution of features to the model performance. In the presence of codependent features, the predictive information is divided among them. For example, information about storm intensity is shared among variables like maximum updraft speed, composite reflectivity, and mid-level updraft helicity. A univariate measurement of importance would ignore those important codependencies and erroneously rank some features too low. For example, for the severe hail dataset, the low-level lapse rate and major axis length features may be ranked higher than the composite reflectivity and updraft features because the latter are correlated with each other (Fig. ~\ref{fig:hail_rankings}). One drawback of the multipass methods, however, is that, as more features are permuted, the importance score relies on more and higher-order dependencies and the latter are often poorly sampled. 
        
        For a high-dimensional dataset with a large number of colinear/correlated features, it is difficult to effectively determine how a feature uniquely contributes to a model. In this study, we demonstrate the concept of \textit{grouped} permutation importance \citep{Gregorutti+etal2017_corr,Konig+etal2020,Au+etal2021} where the goal is to highlight how a grouping of two or more features contributes to the performance of the model. Groupings can be determined automatically through feature clustering based on correlations, but for improved explainability, it is useful to manually select the feature groups (e.g., storm vs. environment features for severe weather; temperature vs. radiative features for road surface temperatures). Another key advantage of the grouped feature importance is the reduced computational time versus other permutation importance methods, assuming  the number of groups $\ll$ number of features. As with backward and forward permutation importance, we can conceptualize two forms of grouped feature importance:
        \begin{itemize}
            \item How does removing this group of features reduce model performance? 
             \item How does the model perform when relying solely on this group of features? 
        \end{itemize}
        For the first method, all features in the group have their samples jointly permuted such that the connection to the target variable is lost, but the joint distribution between the features is preserved, whereas for the second method, all features not in the group are jointly permuted. Similarly to \citet{Au+etal2021}, we refer to these two methods as the Grouped and Grouped Only permutation importance, respectively.  

        According to \citet{InterpretMLTextbook}, favoring the training or testing dataset for feature importance remains an open question (see their section 5.5.2).  \citet{Lakshmanan+etal2015}, however, argued for only using the training dataset. The goal of measuring feature importance is to quantify how the model relies on each feature and not how well the model generalizes to unseen data. If the ML model learned a pattern in the training dataset that is underrepresented in the independent dataset, then evaluating feature importance on the testing dataset can bias our understanding of how the model works. For example, the training dataset could have temperature ranges of $-15 \text{--} -10^{\circ}$C while the testing dataset range is largely -5-5$^{\circ}$C. If the ML model heavily relied on temperatures $<$ -10 $^{\circ}$C to predict freezing road surfaces, we would fail to determine that using the testing dataset. One could evaluate importance on both the training and testing to identify any discrepancies, but then you'd have to ascertain whether the difference is due to poor sampling or overfitting. Therefore, the feature importances and the various experiments in this study are evaluated using the training dataset.

        We use the normalized area under the performance diagram curve (NAUPDC; \citealt{Flora+etal2021, Miller+etal2021}) as our performance metric. The area under the performance diagram curve (AUPDC) is a summary of the performance diagram curve \citep{Roebber2009}, which measures how well the model discriminates between correct predictions and false predictions while ignoring correct negatives. The normalization is to account for the sensitivity of AUPDC to the base rate ( \citealt{Boyd+etal2012})\footnote{base rate is referred to as skew in \citet{Boyd+etal2012}}. To improve the ranking estimates, we recompute the rankings through several permutations (N=30) and compute the mean feature importances. Using multiple permutations limits the influence of random permutations, which will occasionally replace a feature value with an inappropriately similar value and, thereby, low-bias the importance estimate.

    \subsubsection{Model-specific Methods}\label{sec:model_specific}
        In this study, we use logistic regression and random forest models, which have built-in feature relevance methods. For logistic regression, we can compute the relevance of feature $x_i$ as 
        \begin{equation}
             R(x_i) = |\beta_i| * std(x_i) 
        \end{equation}
        where $\beta_i$ is the coefficient of feature $x_i$ and $std(x_i)$ is the standard deviation of $x_i$. Since we standardize the logistic regression input, each feature ideally has a unit standard deviation, so the relevance is theoretically based only on $|\beta_i|$. For random forests, we can use the impurity importance \citep{Breiman2001_RF, McGovern+etal2019_blackbox}, which ranks the features according to how many examples they affect (e.g., whether they occur higher in the tree) and how effectively they split the data (i.e., increase the information gain). Unfortunately, this method is biased, as it will tend to rank features with a higher cardinality (more unique values) higher than other features and is sensitive to correlated features \citep{Strobl+etal2007, Strobl+etal2008}. 

    \subsubsection{Partial Dependence}\label{sec:pd}
        To complement the feature ranking, it is crucial to understand why particular features are important and what their expected contribution is to the prediction of the ML model. A common approach to visualizing feature effects is the partial dependence (PD) plot \citep{Friedman2001, McGovern+etal2019_blackbox, Jergensen+etal2020}. The PD of the feature $x_{j}$ is defined as:
        \begin{equation}\label{eqn:pd}
            PD (x_{j}) = \frac{1}{N}\sum_{i=1}^N f(x_j, \mathbf{X}^{(i)}_{\setminus x_j})
        \end{equation} 
        where $f$ is the ML model, $N$ is the number of training examples, $\mathbf{X}^{(i)}_{\setminus x_j}$ is the set of features excluding $x_j$ and their values for the $i$th training example. The idea is to set all examples for feature $x_j$ to a single value, average the ML predictions, and repeat the process for multiple values of $x_j$ to produce a curve. We then compute the "centered" partial dependence by subtracting the average partial dependence value so that the mean effect is zero:
        \begin{equation}\label{eqn:centered_pd}
            PD_{centered}(x_j) = PD(x_j) - \frac{1}{V}\sum_{v=1}^{V} PD(x_{j,v})
        \end{equation}
        where $V$ is the number of values for which $PD$ was computed for. 

    \subsubsection{Accumulated Local Effects}\label{sec:ale}
        Though the PD curves are easy to calculate and understand, they assume that the features are independent (correlated features can distort the PD curve; \citealt{Molnar+etal2021}) and their marginal effect can hide heterogeneous effects \citep{InterpretMLTextbook}. An alternative to PD is a recently developed method known as accumulated local effects (ALE; \citealt{Apley+etal2016}). The ALE for the feature $x_j$ is:
        \begin{equation}
            ALE_j(x_j) = \int_{\min(z_j)}^{x_j} \mathbb{E} \Bigg[\frac{\partial f(\mathbf{X})}{\partial X_j} \Big| X_j = z_j \Bigg] dz_j - c
        \end{equation}
        where $f$ is the ML model, $\mathbf{X}$ is the set of all features, $z_j$ are the values of $x_j$, and $c$ is the integration constant. The constant $c$ is chosen as the mean of $ALE(x_j)$, so the mean effect is zero, similar to Eqn.~\ref{eqn:centered_pd}.

        For a given feature, ALE computes the expected change in prediction over a series of conditional distributions and then accumulates (integrates) them to return the feature effect. By computing the average change in prediction over a series of small windows, ALE isolates the effect of the feature from the effects of all other features and avoids the pitfall of partial dependence, which can suffer from unlikely or non-physical combinations of feature values. More details on the ALE calculation are provided in \citet{InterpretMLTextbook} and \citet{Flora2020}. 

        Recently, \citet{Greenwell+etal2018} introduced a simple method that ranks features based on the variance in their PD curve. In this study, we use the ALE curve instead of PD as we deal with multicollinear features. To measure the variance, we calculate the standard deviation of the first-order ALE for each feature. Whether using ALE or PD, the method assumes that the greater the variation in the curve, the more relevant the feature. 

\subsection{Local Explainability Methods}\label{sec:local_explain}
    Though global explainability methods can provide a good overview of the model, the explanation of a single prediction can differ from the global explanation. A common technique to explain individual predictions (or some small subset thereof) is to approximate the final prediction [$g(x)$ is approximate of $f(x)$] as the sum of individual contributions ($\phi_i$) from each feature:
    \begin{equation}\label{eqn:sum}
        f(x) \approx g(x') = \phi_0 + \sum_{i=1}^N \phi_ix_i'
    \end{equation}
    where $\phi_0$ is the bias (often the average prediction of model $f$), $N$ is the number of features and $x' \in \{0,1\}^M$ (indicating whether a feature is contributing or not). Methods to determine these individual contributions are known as feature attribution methods. In the following sections, we briefly highlight three such methods. 
    One goal of this paper is to determine whether local explainability methods can be scaled to provide global explanations. To capture global characteristics, we calculated the feature attributions values from 50,000 random samples extracted from each dataset. To assign a feature relevance score, we sum the absolute feature attribution values of each example per feature. To compute the feature effects, we compute the average feature attribution value per bin (using the same bins as for the ALE and PD curves).
    
    \subsubsection{Tree Interpreter}\label{sec:ti}
    \begin{figure}[t]
        \noindent\includegraphics[height=26pc, width=22pc,angle=0]{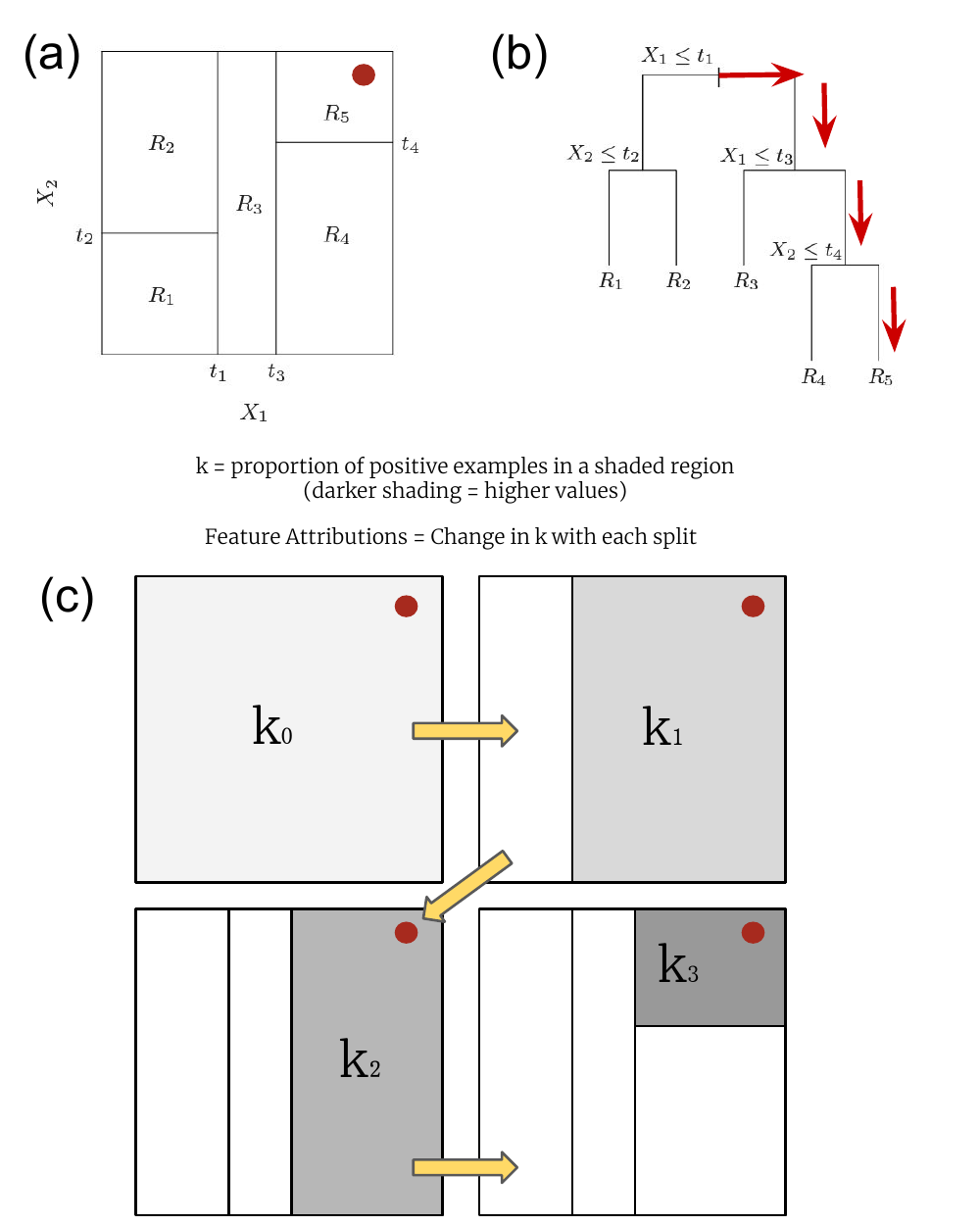}
        \caption{ Illustration of the tree interpreter method for a simple decision tree (b). The 2-feature domain in (a) is split according to the decision tree in (b). The data point (red) in $R_5$ indicates the example to be explained. $k_i$ is the proportion of positive examples in the highlighted region. 
        }
        \label{fig:tree_interpret}
    \end{figure}
        Tree interpreter is a feature attribution method for decision tree-based models \citep{Saabas2014, Loken+etal2022}. For a classification decision tree with $M$ leaf nodes, the feature space is divided into $M$ regions. The prediction is then defined as follows:
        \begin{equation}
            f(x) = \sum_{m=1}^M c_m I(x, R_m)
        \end{equation}
        where $R_m$ is the $m$th region, $c_m$ is the proportion of positive examples in $R_m$ (from the training dataset), and $I$ is the indicator function (return 1 if x $\in$ $R_m$, 0, otherwise). When producing a prediction from a decision tree, we can visualize the path the data take through the various branches until they reach a leaf node. For example, Figure ~\ref{fig:tree_interpret} has a 2-feature domain split into 5 regions (Fig.~\ref{fig:tree_interpret}a) based on the decision tree in Fig.~\ref{fig:tree_interpret}b. To compute the feature attributions for the single data point (shown in red), we progressively parse the domain as we pass through the nodes of the decision tree. Initially, the data are not split and $k_o$ is the total number of positive class examples divided by the total number of examples. Once we divide the data according to $X_1 \leq t_1$, the proportion of positive examples has changed and is represented by $k_1$. The change from $k_0$ to $k_1$ is attributed to the feature $X_1$. With each additional split, there are additional feature attributions for $X_1$ or $X_2$. In terms of equation ~\ref{eqn:sum}, $\phi_0 = k_0$ and $\phi_i = \sum \Delta k^{i}$ where $\sum \Delta k^{i}$ is the sum of changes to $k$ associated with the $i$th feature. For a random forest, we can determine the contribution breakdown for each tree and then average the results together. 
         
    \subsubsection{Local Interpretable Model-Agnostic Explanations}\label{sec:lime}
        Linear regression models are highly interpretable models (see Fig.~\ref{fig:int_vs._exp}), and we can leverage their interpretability by using them as local surrogates for more complex models to explain individual predictions. The local interpretable model-agnostic explanation (LIME; \citealt{Ribeiro+etal2016}) fits a weighted linear regression model on perturbed data in the neighborhood of the example to be explained. The input features in the local model are binary representations of themselves such that the coefficients are $\phi_i$ and follow equation ~\ref{eqn:sum}. To determine the neighborhood, LIME uses an exponential smoothing kernel. One drawback of this method is the need to determine the properties of this kernel (e.g., its width). We use the default kernel width of 0.75 times the square root of the number of features, a sample size of 2500, and compute the coefficients for all features.  
        
    \subsubsection{Shapely-based Methods}\label{sec:shap}
        Shapley values \citep{Shapley1953}, which are rooted in game theory, have become one of the most promising methods for explaining ML predictions, as they can be useful for both local and global explainability. The Shapley value for the feature $x_j$ is the weighted average difference in model prediction (its contribution) when $x_j$ is included and not included in a feature subset $S$ for all possible $S$. From a game theory perspective, when players are cooperating in a coalition, Shapley values are the fairest possible payouts to the players depending on their contribution to the total payout. In terms of ML, we can think of players as features and payouts as their contributions to the final prediction (the total payout). A more complete description of the Shapley values is provided in \citet{InterpretMLTextbook}. 

        One limitation of traditional Shapely values is that they treat features independently, ignoring correlated or multicollinear features. To address this, we can group the features into coalitions and compute an extension of the Shapely values known as Owen values \citep{Owen1977,Lopez+Saboya2009}. To compute the Owen value for $x_j$, we compute the weighted average change in model prediction when $x_j$ is included and not included in all possible feature subsets, but such that the subsets exclude features from one grouping. We repeat the calculation with each feature grouping being excluded, and average those values. Another key benefit of the Owen values is that the number of feature subsets to evaluate is greatly reduced, especially for a smaller set of feature groupings. The Owen values are calculated using the Shapely Additive Explanation (SHAP) Python package \citep{Lundberg+Lee2016} and we use feature correlations to determine the feature groups (using SHAP's partition masker). 
    
        For all feature attributions, we would like the following axioms to be satisfied. 
        \begin{enumerate}
            \item Local accuracy (additivity): The sum of the contributions of each feature plus the average prediction of the model must be equal to the final prediction for a given instance (Eqn.~\ref{eqn:sum}). 
    
            \item Consistency (monotonicity): If an ML model changes so that the marginal contribution of a feature increases or stays the same, the feature attribution must also increase or stay the same, respectively. 
    
            \item Missingness: Missing features (e.g., features that have been marginalized out) must have a contribution of zero to the model.
        \end{enumerate}
        Of the three feature attribution methods discussed above, the Shapley / Owen values are the only method that satisfies all three of these properties \citep{Shapley1953, Young1985, lundberg2018consistent, Lundberg+etal2020_localtoglobal}. Using an approximate linear model, LIME cannot guarantee the principle of local accuracy, whereas the Tree interpreter satisfies the local accuracy principle but has consistency issues as features near the root can incorrectly be given less weight \citep{Lundberg+etal2020_localtoglobal}. Although the tree interpreter and LIME do not completely satisfy all three principles, in terms of run time (time to generate an explanation for a new sample), LIME and the tree interpreter are considerably faster \citep{Lundberg+etal2020_localtoglobal}, and therefore we consider them for this study. 

        Recently, \citet{Covert+etal2020} used the concept of additive attributions and extended Shapely values to global explainability. The Shapely Additive Global importancE (SAGE) method explains how each feature contributes to the expected model loss rather than to an individual prediction. SAGE is a global and model-agnostic extension of the LossSHAP method \citep{Lundberg+etal2020_localtoglobal}, which computes feature attributions for local loss, but only for tree-based models. One drawback for SAGE is that for missing features it relies on a marginal rather than conditional distribution approach, which is undesirable for datasets with correlated or dependent features.  \citet{Covert+etal2020} note that by using marginal distributions, though, it ensures that features not used by the model receive zero attribution. To compute the SAGE values, we are using the SAGE python package \citep{sage} with default settings.

\section{Results}\label{sec:results}

    \subsection{Global Feature Rankings} 
        \begin{figure*}[t]
            \noindent\includegraphics[width=38pc,angle=0]{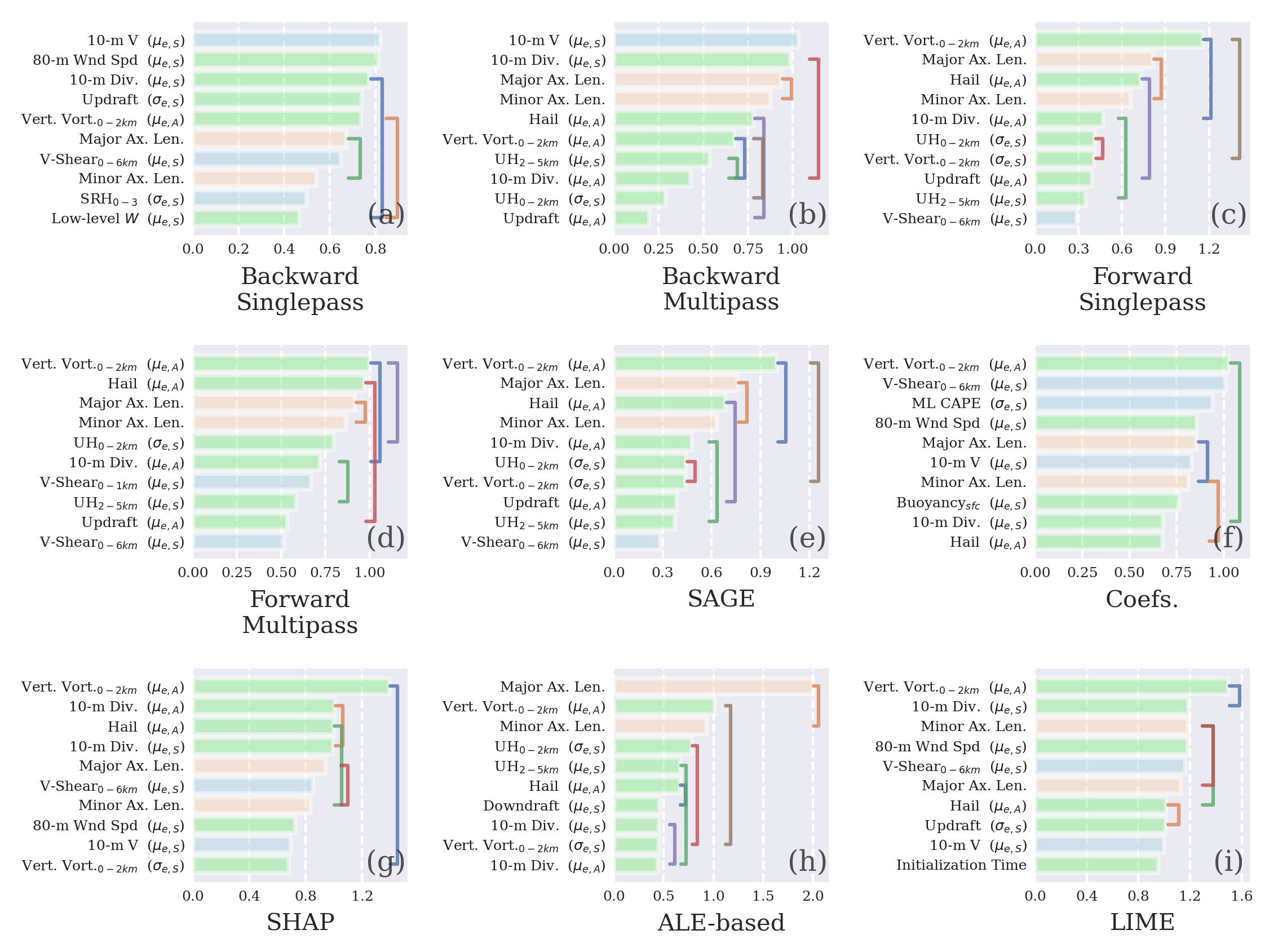}
            \caption{ Feature rankings for the WoFS tornado dataset. Feature importance scores have been scaled by their maximum value to allow for comparisons between methods. Green, blue, and light gold indicate intra-storm, environment, and storm morphological features, respectively. Features with a linear correlation $>$ 0.5 are linked together (each linkage is shown in a different color to help distinguish them from one another). See Table 1 in \citet{Flora+etal2021} for the feature naming convention. }
            \label{fig:torn_rankings}
        \end{figure*}

        Fig.~\ref{fig:torn_rankings} compares the feature rankings for the top 10 features of the WoFS tornado dataset.  There is some consensus on the main features, but there are examples of significant disagreements. For example, the v-component of the 10-m wind is the top feature for both BSP and BMP (which is by default as the first step of multipass is the singlepass result), the ALE variance has major axis length as the top feature (cf. Fig.~\ref{fig:torn_rankings}g), and the remaining methods have 0-2 km vertical vorticity as the top feature. Although there is disagreement between the feature ranking methods, each ranking is physically plausible. Thus, relying on a single explanation could lead to confirmation bias when assessing feature significance \citep{Molnar+etal2021, Ghassemi+etal2021}. For example, \citet{Satyapriya+etal2022_disagree} found that a third of their participants (out of 25) tended to favor an explanation if it better matched their intuition. In Part II, we explore the validity of the different ranking methods and find that the variance in the feature rankings decreases somewhat when excluding methods that poorly assign importance (e.g., assigning higher importance to features that weakly contribute to model performance).

        To summarize the disagreement among the feature ranking methods, we use the top feature and feature rank agreement statistics from \citet{Satyapriya+etal2022_disagree}. The top feature agreement measures the fraction of common features between two top $N$ feature rankings, while the feature rank agreement measures the fraction of top-$N$ features with identical rankings. For this study, instead of requiring identical ranks for a given feature, we count situations where the rank of a feature is off by one between two rankings (e.g., feature $x_i$ is ranked third in one ranking, but fourth in another). Requiring an exact rank match between two methods is likely to overestimate disagreement, especially when tiny differences can separate ranks (e.g., when the number of similarly important features increases) and the explanation of model behavior is nearly identical. 
        
\begin{figure*}[t]
            \noindent\includegraphics[width=38pc,angle=0]{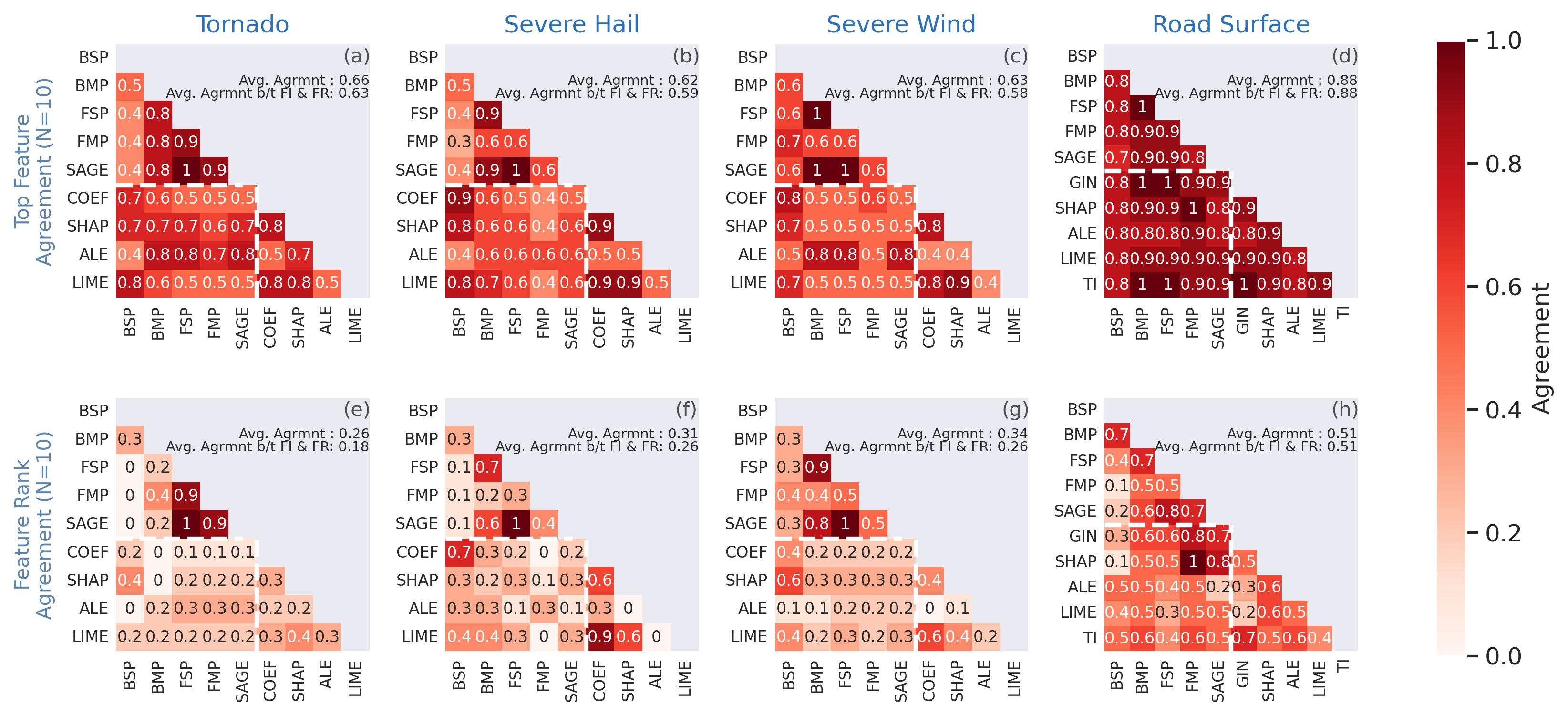}
            \caption{ The agreement between top 10 features (top row) and rank of the top features (bottom row) for the tornado (first column), severe hail (second column), severe wind (third column), and road surface (last column) datasets. The following methods are shown: backward single-pass (BSP), backward multipass (BMP), forward single-pass (FSP), forward multipass (FMP), SAGE, logistic regression coefficients (COEF), SHAP, ALE, LIME, and tree interpreter (TI). TI is a model-specific method, so it is only shown for the random forest model trained on the road surface dataset. Higher values indicate higher agreement. The region showing the agreement between the feature importance (FI) and feature relevance (FR) methods is delineated by white dashed lines. The average agreement between all methods (Avg. Agrmnt) and the average agreement in the delineated region (Avg. Agrmnt b/t FI $\&$ FR) are provided in the upper right corner of each plot.  }
         \label{fig:disagree}
        \end{figure*}

        Fig.~\ref{fig:disagree} shows the top 10 feature and feature rank agreement, respectively, between each pair of feature ranking methods. In general, there is fairly high agreement on the top 10 features for the road surface datasets (Fig.~\ref{fig:disagree}d) whereas there is more modest agreement for the WoFS datasets (Fig.~\ref{fig:disagree}a-c). 
        
        To some extent, disagreement between the explainability methods is anticipated, as different explainability methods are designed for different tasks. This is a concept that is echoed in forecast verification, where different metrics are designed to measure different aspects of forecast quality \citep{Murphy1993a}. However, even when designed for a particular task, disagreements between explainability methods can arise due to the impact of correlated features \citep{Gregorutti+etal2017_corr, Strobl+etal2007, InterpretMLTextbook}, bringing into question the validity of their feature rankings. For example, when there are correlations among informative features, their single-pass permutation importance can be reduced, and depending on the strength of the correlations, the features less correlated with the target variable may appear more important \citep{Gregorutti+etal2017_corr}. There is decent agreement on the top 10 features between the BSP and the logistic regression coefficients (Fig.~\ref{fig:disagree}a-d), which, when the features are independent, should provide identical feature rankings \citep{Gregorutti+etal2017_corr, Hooker+etal2019}. Similarly, there is high agreement between SHAP and the logistic regression coefficients (Fig.~\ref{fig:disagree}a-d), another pair of methods that should provide identical rankings given independent features. This is consistent with the facts that the logistic regression model uses L1 norm regularization and the SHAP values are based on Owen values, both of which mitigate the effects of the correlations between features. ALE and SAGE generally agree most with the forward-pass methods, which is consistent with the fact that forward methods largely ignore feature interactions (since they begin with all features permuted), and the ALE variance is determined by first-order effects and SAGE assumes feature independence. The local attributions (LIME, TI, and SHAP), tend to have higher agreement for all four datasets, but only have modest agreement with the feature importance methods. This result suggests that local attributions can scale as global feature relevance methods, but may not translate as global feature importance methods. 
        
        Explainability methods can disagree by design, but it is unclear whether the disagreement is inflated due to the inclusion of poorer performing methods (i.e., assigning higher importance to a feature that contributes weakly to model performance). It is possible that by evaluating a large set of feature ranking methods disagreement is likely to be higher. However, the disagreement among all the methods is similar to the average disagreement between the feature importance and feature relevance methods. This similarity suggests that the disagreement is not solely based on including several feature ranking methods or due to the explainability methods being designed for different tasks or including several methods, but is partly attributable to differing accuracy among the methods (explored in Part II).

        In terms of rank agreement, there is more agreement on the rank for the severe wind and road surface datasets (Fig.~\ref{fig:disagree}g-h) than for the tornado and severe hail datasets (Fig.~\ref{fig:disagree}e-f). In Part II, we find that the severe wind and road surface datasets have weaker feature interactions than the tornado and severe hail datasets (see Section 6a of Part II), which means that the models rely more heavily on first-order effects. By relying more on first-order effects, the feature importance estimate is improved since the importance varies less across the feature space (i.e., more agreement between local and global explainability). As mentioned above, the logistic regression coefficients and the BSP methods should provide identical rankings if the correlations do not strongly impact the identification of the top 10 features. However, there was little rank agreement between the two methods (Fig.~\ref{fig:disagree}e-h) indicating a strong influence of correlated features. An outlier is the perfect agreement between SAGE and FSP for the WoFS datasets (Fig.~\ref{fig:disagree}e-g). Recall that FSP measures the importance of a feature when all other features are permuted, which breaks up any feature interactions or correlations. The similarity between SAGE and FSP suggests that SAGE may be ignoring feature interactions, but further research is required to test this hypothesis. Lastly, the TI method had the greatest rank agreement with Gini importance (GIN, 0.7), which can poorly assign rank when features are correlated \citep{Strobl+etal2007, Strobl+etal2008}. We know that this limitation is by-product of how random forests are trained \citep{Strobl+etal2008}, which is likely negatively impacting TI as a model-specific explainability method.  
        
        As noted in Section \ref{sec:explain_methods}\ref{sec:global_explain}.\ref{sec:feature_rankings}.\ref{sec:flavors_of_perm} and given the disagreement between explanation methods, it becomes increasingly difficult to effectively distinguish how a feature uniquely contributes to model performance when it is colinear/correlated with other features. Thus, we demonstrate the concept of \textit{grouped} permutation importance where the goal is to highlight how a grouping of two or more features contributes to the performance of the model. For the WoFS-based datasets, we separated the features into one of three groups: intrastorm, environmental, and storm morphology. For the road surface dataset, we created temperature- and radiation-based groups. 

\begin{figure*}[t]
            \noindent\includegraphics[width=38pc,angle=0]{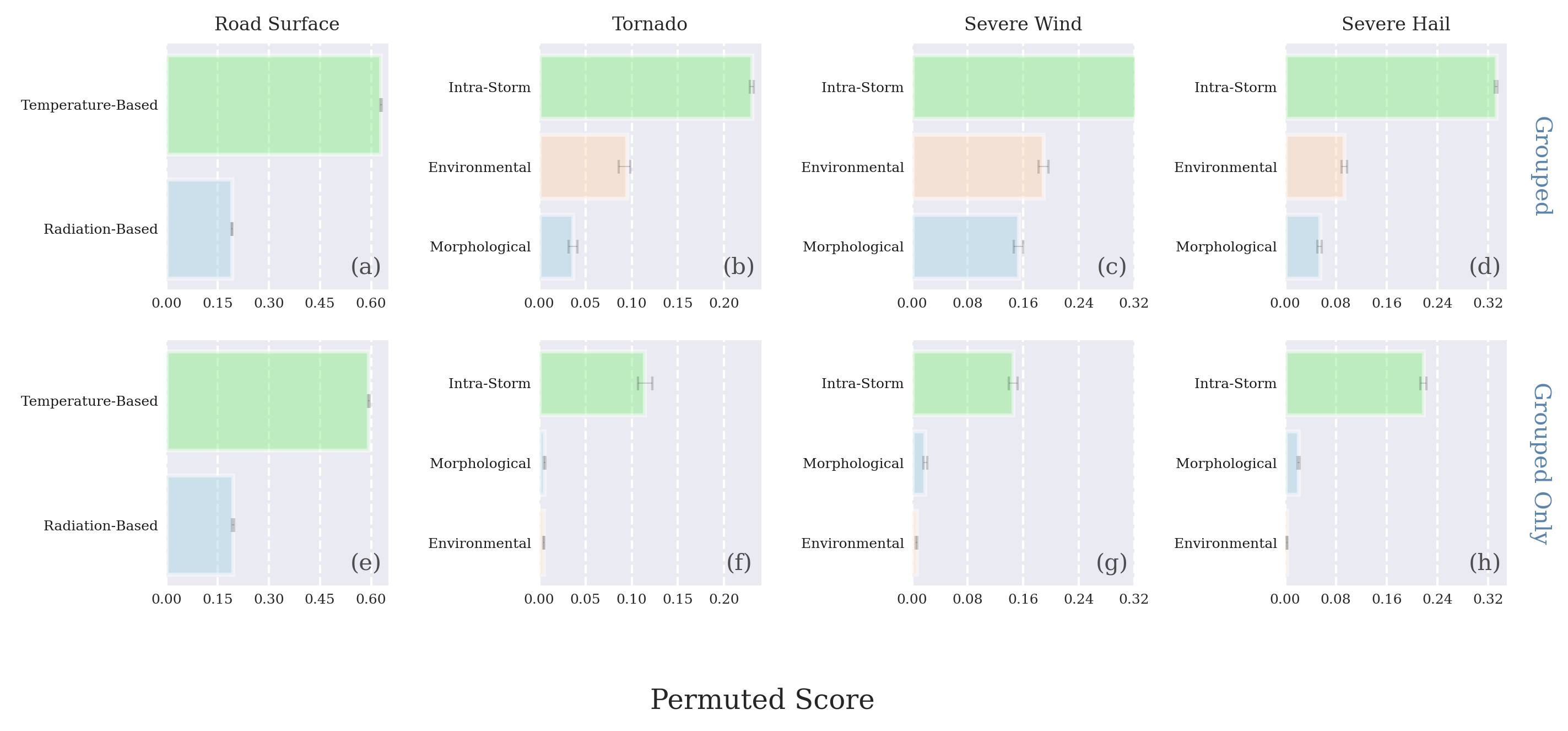}
            \caption{ Feature rankings based on the Grouped (top row) and Grouped Only (bottom row) methods for the original road surface (first column), tornado (second column), severe wind (third column), and severe hail (last column) datasets. Grouped importance assesses how removing a group of features (jointly permuting them) reduces model performance while Grouped Only assesses how the model performs when solely relying on a group of features (i.e., all other features are jointly permuted). For the severe weather datasets, color-coding same as Fig.~\ref{fig:torn_rankings}. For the road surface dataset, temperature-based are shown in green and radiation-based are shown in blue. }
             \label{fig:grouped_rankings}
        \end{figure*}

        For the WoFS datasets, intrastorm features were the most important, especially for the Grouped Only results (Fig.~\ref{fig:grouped_rankings}), which is not surprising given that storm-scale information remains predictable at these lead times \citep{Flora+etal2018} and the intrastorm features are the most causally related to storm hazards. We do not conclude that environmental and storm morphological features are unimportant but rather that, by themselves, they are not nearly as predictive as the intrastorm features are by themselves. For example, the importance of the environmental features is much higher for the Grouped than Grouped Only, indicating that if they are missing the model performance suffers (Fig.~\ref{fig:grouped_rankings}b,c,d), but without the other features they contribute little to model skill (Fig.~\ref{fig:grouped_rankings}f,g,h). This result aligns with another ML study that leveraged the WoFS dataset where intra-storm predictors were found to be more important than environmental predictors \citep{Clark+Loken2022}. 
        
        For the road surface dataset, it is not surprising that temperature-based features are more valuable to the ML model than radiation-based features. However, unlike the WoFS dataset, there is little difference between the Grouped and Grouped Only results, suggesting that the road surface model is strongly dependent on the temperature-based features, which largely agrees with the individual importances (see Fig. 4g in Part II). Note that the two groups in the road surface dataset do not mutually account for all features. If they had, then Grouped and Grouped Only results would be identical by default. 
        
    \subsection{Global Feature Effects}\label{sec:effect_results}
        To improve our understanding of the feature rankings and the ML models, we can estimate the feature effects. For this study, we compare feature effects estimated by PD, ALE, SHAP, LIME, and tree interpreter (TI). To aid in the interpretation of the feature effects, we compute the conditional event rate per feature [i.e., $p(y=1 | x_i)$]. To compute the conditional event rate, we use a Bayesian histogram method (Python package bayeshist; \citealt{Hafner_bayeshist}). This method assumes a beta distributed prior [$p(y=1) \approx \mathcal{B}(\alpha, \beta)$ where $\alpha$ and $\beta$ are shape parameters] and a similar distribution for the posterior [$p(y=1|n_i^+, n_i^-) \approx \mathcal{B}(\alpha+n_i^+, \beta+n_i^-)$] where $n_i^+, n_i^-$ are the number of positive and negative samples in the $i$th bin, respectively. By binning feature $x_i$'s values, we can compute $\mathcal{B}$ in a series of quantiles. To keep only significant bins, the method compares each pair of neighboring bins and combines the pair if they are likely from the same event rate sample. 
        \begin{figure*}[t]
         \noindent\includegraphics[width=38pc,angle=0]{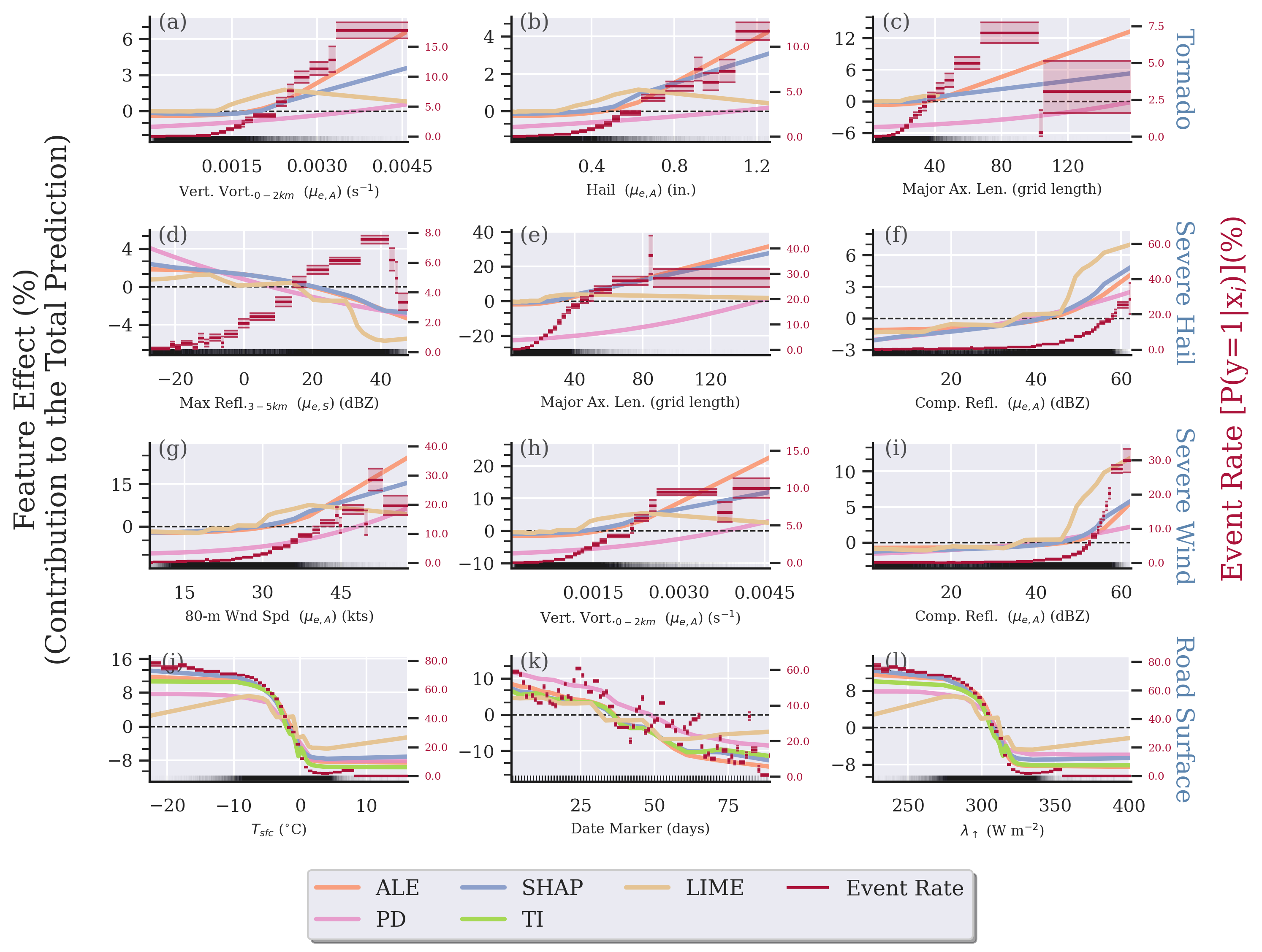}
            \caption{ Comparison of the the first-order feature effects derived from PD (pink), ALE (orange), SHAP (blue), LIME (gold), tree interpreter (TI; green) with the event rate (red) for the top 3 features for the tornado (first row), severe hail (second row), severe wind (third row), and road surface (last row) datasets. TI is a decision-tree-specific method and therefore is only applicable to the random forest trained on the road surface dataset. The red shaded regions indicate the 95$\%$ confidence intervals for the event rate. Black tick marks along the x-axis indicate the marginal distribution for each feature. }
            \label{fig:feature_effects}
        \end{figure*}
        
        The measured feature effects and the event rate curve are shown for the top three features of each dataset (Fig. ~\ref{fig:feature_effects}). The top features were determined based on the median feature rankings based on all methods (this ranking can be seen in Fig. 4 in Part II). The general similarity between the feature effect curves and the conditional event rate suggests that the ML models are learning appropriate relationships. By comparing against the conditional event rate we can better determine whether the learned relationships match physical intuition, which can lead to improved trust in the model. For example, higher vertical vorticity corresponds to a higher probability of a tornado (Fig. ~\ref{fig:feature_effects}a) and lower surface temperature ($T_{sfc}$) corresponds to a higher probability of a road surface freezing (Fig. ~\ref{fig:feature_effects}j).  One notable outlier is the 3-5 km maximum reflectivity (Fig. ~\ref{fig:feature_effects}d) where the learned relationship is opposite to the event rate trend. Composite reflectivity (Fig. ~\ref{fig:feature_effects}f), which is highly correlated with 3-5 km maximum reflectivity (not shown), has a better learned relationship to the prediction. We attribute these results to the fact that when two features are highly correlated, regression models can learn opposite effects for the two features with the feature that is the stronger predictor of the target variable having the "correct" effect. These results highlight the inherent difficulties of applying explainability methods to models with physically correlated data and motivate the use of dimensionality reduction approaches to reduce the impact of correlated features (see Part II).

        In terms of the correspondence between methods, ALE and SHAP tend to agree well with each other and with the event rate, while PD seriously underpredicts some feature effects. PD can have difficulty estimating the mean effect due to its inability to distinguish between the effects of correlated features.  The correspondence of LIME with other methods and the event rate varies by dataset and feature. For example, LIME is similar to the other curves for the road surface dataset 
        (Fig. ~\ref{fig:feature_effects}j-l), but has weaker correspondence for higher values of vertical vorticity, hail, and 80-m wind speed in the tornado dataset (Fig. ~\ref{fig:feature_effects}a-c).  While ALE is designed to measure the global feature effects, SHAP is a local explainability method; the similarity between SHAP and the event rate suggests that SHAP can scale well (i.e., summarizing local explanations leads to useful global explanations). ALE and SHAP have modest disagreements on some of the features (e.g., Fig~\ref{fig:feature_effects}a,  Fig~\ref{fig:feature_effects}b, Fig~\ref{fig:feature_effects}i). For example, for higher values of 0-2 km vertical vorticity ALE has a higher estimated effect than SHAP, which corresponds with the higher base rate for higher vorticity values suggesting that SHAP is underestimating the global feature effect. In general, the correspondence between the methods seems largely dependent on sample size, where smaller sample sizes are associated with larger feature effect disagreement. 
        
        \begin{figure*}[t]
            \centering
            \noindent\includegraphics[width=30pc,angle=0]{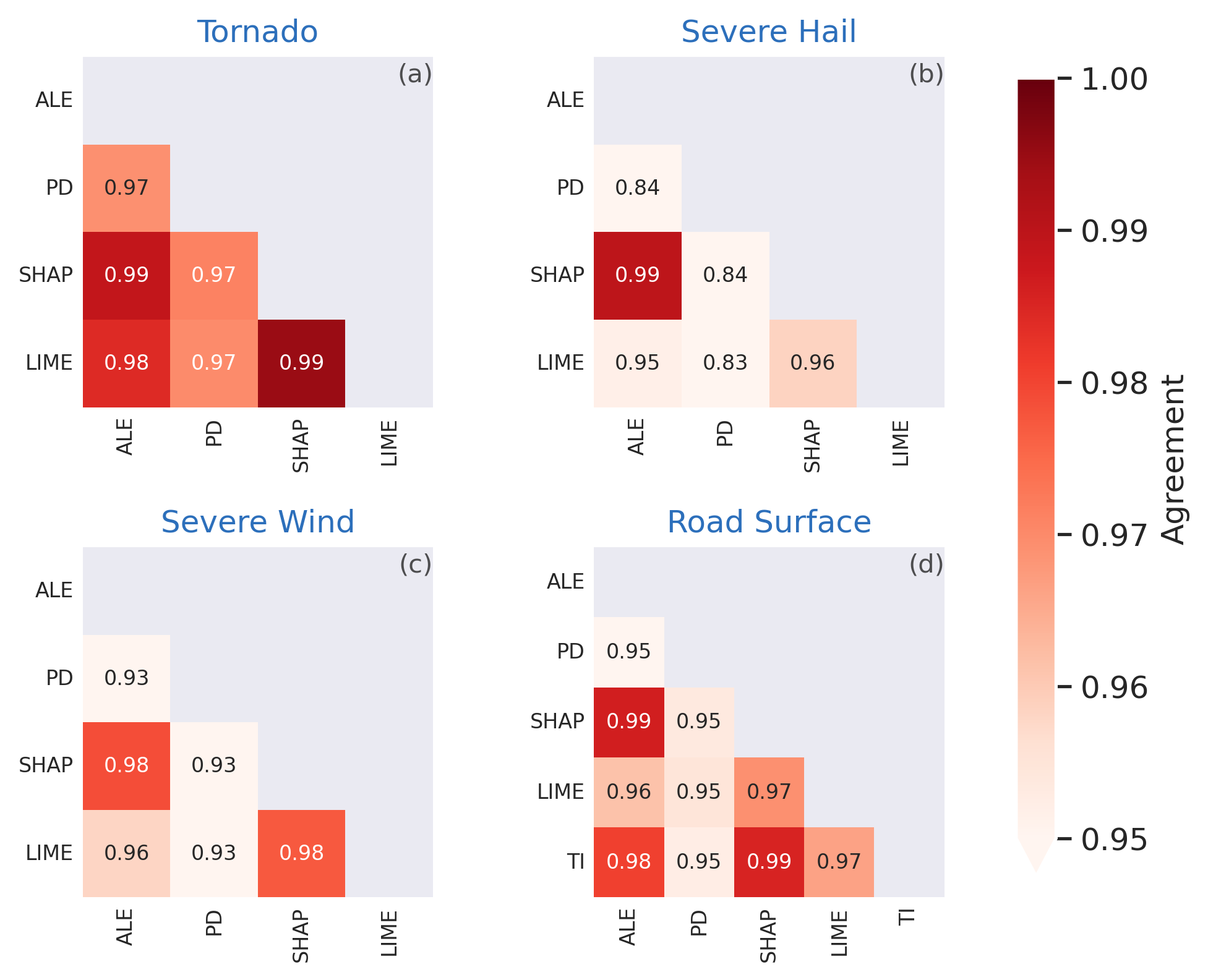}
            \caption{ The agreement (1-RMSD) between feature effect curves from pairs of methods for the (a) tornado , (b) severe hail, (c) severe wind, and (d) road surface datasets. TI is a model-specific method, so it is only shown for the random forest model trained on the road surface dataset. Higher values indicate higher agreement. }
         \label{fig:disagree_effects}
        \end{figure*}
        
        To summarize the disagreement between two methods, we calculated the root mean squared difference (RMSD) for all features and computed the variance-weighted average (variance in the feature effect curve). To show the differences in terms of agreement, Fig.~\ref{fig:disagree_effects} is based on 1\text{--}RMSD. Unlike the feature ranking agreement (Fig.~\ref{fig:disagree}), there is fairly strong agreement amongst the different methods on feature effects. This result aligns with our subjective impression of Fig.~\ref{fig:feature_effects} that the curves correspond well with each other when the sample size is large enough. ALE, SHAP, LIME, and TI tend to agree more with each other, while PD has lower agreement with the other methods. However, PD is more consistent with other methods in the road surface datasets than in severe weather datasets (cf. Fig.~\ref{fig:disagree_effects}d and Fig.~\ref{fig:disagree_effects}a-c). The feature correlations in the severe weather datasets are stronger and more frequent than in the road surface dataset (correlation diagrams are shown in Figs. 1 and 2 of Part II), which possibly explains why PD is inconsistent with the other methods in the severe weather datasets.

\section{Practical Advice on Model Explainability}\label{sec:advice}
    We have demonstrated that ML explanation methods can substantially disagree with one another in terms of feature rankings and feature effects. A significant source of disagreement is due to correlated features as distinguishing individual importance/relevance becomes an involved process. Despite these disagreements, our attempts to understand our ML models are not in vain; rather, we should approach this task with certain expectations. For example, for the datasets and models used in this study, the different feature ranking methods were in reasonable agreement on the top 10 features, but disagreed on exact rankings (Fig.~\ref{fig:disagree}). If we de-emphasize interpreting exact rank, then by using multiple ranking methods, we can increase our confidence about the top contributors to our model. One such approach is demonstrated in Part II where we compute the median feature ranking from all methods and display the IQR as a measure of ranking uncertainty. By incorporating uncertainty information, we can improve our confidence and trust in the output from the various explainability methods. 

\begin{figure}[t]
            \noindent\includegraphics[width=20pc,angle=0]{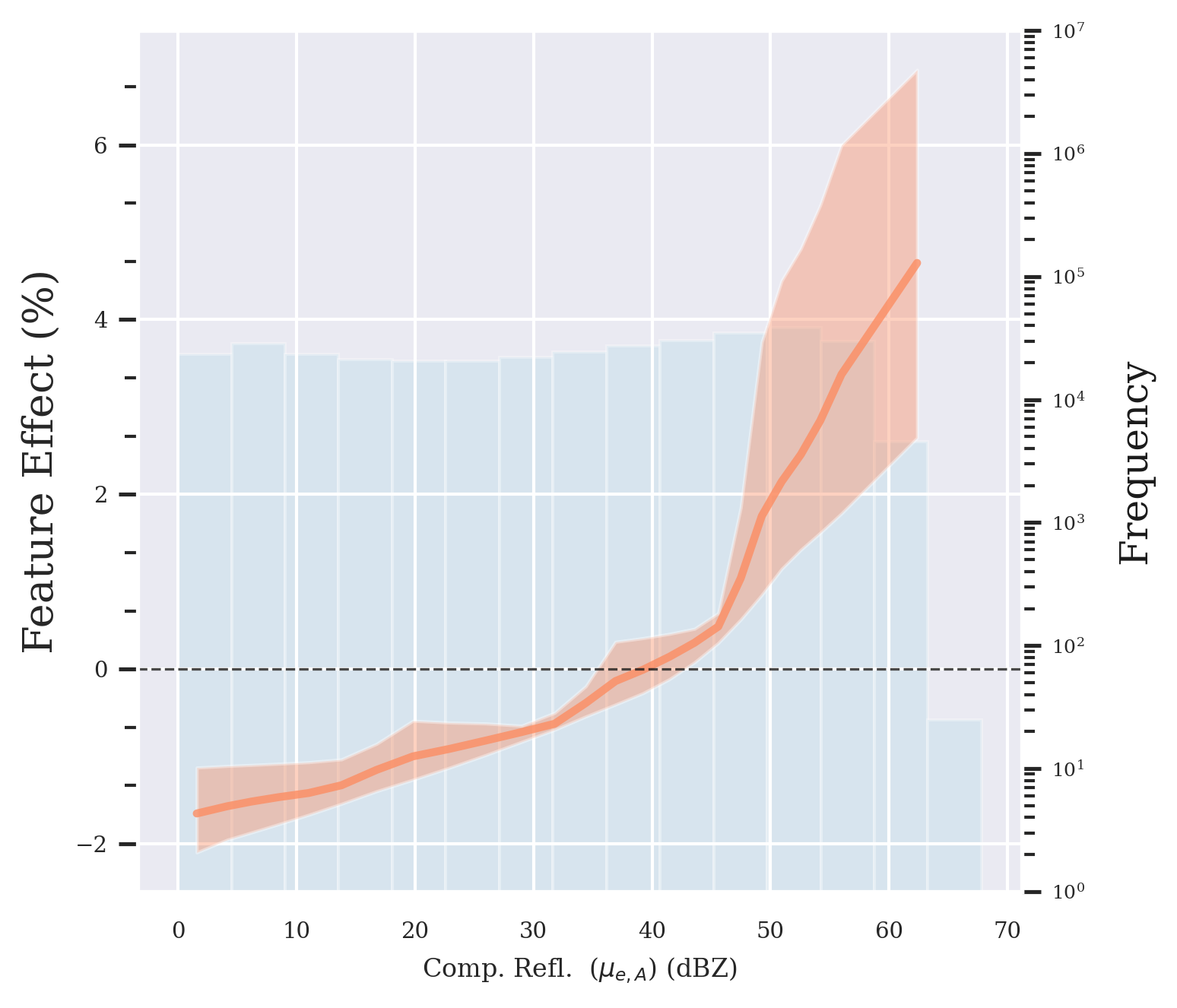}
            \caption{  The method-average (SHAP, LIME, PD, and ALE) feature effect for composite reflectivity in the severe hail model. The composite reflectivity distribution of the training distribution is shown in light blue. }
             \label{fig:avg_effects}
        \end{figure}

    In most cases, the different feature effect methods had high agreement (Fig.~\ref{fig:feature_effects} and Fig.~\ref{fig:disagree_effects}). Disagreements were largely confined to poorly sampled regions where we should be especially cautious about interpreting any explanation method. Fig.~\ref{fig:avg_effects} shows the method-average feature effect for composite reflectivity for the severe hail model. There is decent consistency amongst the methods for low and moderate composite reflectivity values, but effect uncertainty increases as the sample size decreases. The uncertainty does start to increase prior to diminishing samples for composite reflectivity, which is related to the limited sample sizes of the other features (not shown). We can still conclude that higher composite reflectivity values increase the likelihood of severe hail, but we must also acknowledge that the magnitude of this effect becomes uncertain at high reflectivity values.   

\section{Conclusions}\label{sec:conclusions}
Motivated by the increasing interest in explaining machine learning models, the first part of this two-part study synthesizes recent research on many explainability methods. This includes distinguishing explainability from interpretability (Fig.~\ref{fig:int_vs._exp}), local versus global explainability (cf. section \ref{sec:explain_methods}\ref{sec:global_explain} and section \ref{sec:explain_methods}\ref{sec:local_explain}), and feature importance versus feature relevance (section \ref{sec:explain_methods}\ref{sec:global_explain}.\ref{sec:feature_rankings}). We demonstrate visualizations of the different explainability methods, how to interpret them, and provide a comprehensive Python package \citep{Flora+Handler} to enable other researchers to use these methods. Summaries of the key ideas for each method are provided in Fig.~\ref{fig:summary} and Table \ref{tab:rankings_methods}. 

The growing number of ML model explainability tools can make it difficult to select the best method to explain a particular dataset or model. We know that no single method can describe all aspects of model explainability, but the degree to which explanations tend to disagree with each other and how much the validity of the different methods varies has not been thoroughly explored.  We have therefore reviewed various explanation methods and analyzed the differences between for global feature rankings and global feature effects. To generalize the results, we applied the explainability methods to two very different meteorological datasets: a convection-allowing model dataset for severe weather prediction, and a nowcasting dataset for sub-freezing road surface prediction. We conclude the following:
\begin{itemize}
\item The feature ranking methods explored in this study tend to agree on the top features ($N$=10), but substantially disagree on specific ranks. Given the large number of correlated features in our data sets, this result suggests that correlations affect the ranking of top features without radically changing the top features set. This hypothesis is further supported by consistently high agreement between backward single-pass permutation importance and logistic regression coefficient-based rankings.

\item The overall disagreement between all the feature ranking methods for the top features was similar to the average disagreement between the feature importance and feature relevance methods. This suggests that the disagreement is not solely attributable to the feature ranking methods being designed for different tasks. 

\item Grouped feature importance \citep{Au+etal2021} can mitigate issues associated with correlated features and may provide clearer interpretations than individual importances. 

\item The different feature effect estimations had high agreement though PD was least consistent with other methods. Method disagreements were largely due to small sample sizes.

\item Like global feature rankings, the feature attribution methods had high agreement on the top features and less agreement on specific ranks. Although the tree interpreter has inconsistency issues \citep{Lundberg+etal2020_localtoglobal}, its similarity to SHAP and its much faster runtime suggest that it should be further explored in future studies. 

\item Tree interpreter (TI) is a relatively unexplored method, but it is being leveraged in the atmospheric science community \citep{Loken+etal2022}. TI has inconsistency issues \citep{Lundberg+etal2020_localtoglobal} and we found it ranked features similar to Gini importance, which has known issues with correlated features \citep{McGovern+etal2019_blackbox}. However, as a global feature effect method, TI had nearly perfect agreement with SHAP. Given these mixed results, we recommend it be explored further. 

\end{itemize}

This study was one of the first to quantify differences in the various explanation methods, but there were limitations.  First, a common criticism of interpretability and explainability is that there is no objective definition of these terms \citep{Molnar+etal2020_imlpaper}. The fields related to improving ML model understanding \textemdash interpretable ML and explainable artificial intelligence \textemdash are rapidly developing and not yet mature, and so it is not surprising that the nomenclature is still evolving. In this paper, we have provided clear definitions of interpretability and explainability, as well as a brief summary of related terms. Although we did not provide quantitative definitions, we believe that such definitions are not required to quantify important aspects of explainability like ground-truth faithfulness, predicted faithfulness, stability and/or fairness \citep{OpenXAI}. In recent research, benchmark datasets where the ground truth is known are being used to evaluate explanation methods for deep learning \citep{Mamalakis2021}, and in Part II and \citet{Covert+etal2020}, experiments are performed to measure the correspondence between feature importance scores and model performance. Second, a key component of model explainability, feature interactions (i.e., the impact on a feature's effect due to the relationship with another feature), was not explored in this study. After establishing the important predictors and their first-order effects, it is useful to assess the existence and strength of the relationships between the predictors. For many real-world problems, we know that the effect of two or more predictors working together can contribute significantly to explaining a phenomenon and the performance of the ML model. Modeling feature interactions is touted as one strength of ML models, but limited work has been done to measure them. \citet{Friedman+Popescu2008} have developed the most comprehensive set of metrics for evaluating feature interactions, including various statistics that describe how much second-order effects depart from the additive effect between two predictors. However, feature interaction explainability methods are often computationally intractable for more than two features and distinguishing the significance of second- or higher-order interactions from noise is prohibitively difficult.  Lastly, we need to improve our understanding of how disagreements between explanation methods impact end user trust and the biases that may develop from a developer or end user selecting certain methods. For example, \citet{Satyapriya+etal2022_disagree} found that in practice, practitioners often rely on heuristics when choosing explainability methods. In light of these limitations of extant research, we recommend that more concerted effort be devoted to improving our understanding of explainability methods, their strengths and limitations, and their applicability. 

\clearpage
\acknowledgments
Funding was provided by NOAA/Office of Oceanic and Atmospheric Research under NOAA-University of Oklahoma Cooperative Agreement $\#$NA21OAR4320204, U.S. Department of Commerce. The authors thank Eric Loken for informally reviewing an early version of the manuscript. We also acknowledge the team members responsible for generating the experimental WoFS output, which include Kent Knopfmeier, Brian Matilla, Thomas Jones, Patrick Skinner, Brett Roberts, and Nusrat Yussouf.  This material is also based upon work supported by the National Science Foundation under Grant No. ICER-2019758. Any opinions, findings, and conclusions or recommendations expressed in this material are those of the author(s) and do not necessarily reflect the views of the National Science Foundation. 


\datastatement
The experimental WoFS ensemble forecast data and road surface dataset used in this study are not currently available in a publicly accessible repository. However, the data and code used to generate the results here are available upon request from the authors. The explainability methods were computed and visualized using the scikit-explain python package \citep{Flora+Handler} developed by Dr. Montgomery Flora and Shawn Handler. The python scripts used to generate the figures in Part I and II are available at https://github.com/monte-flora/compare-explain-methods. 

\bibliographystyle{ametsocV6}
\bibliography{references}

\end{document}